\documentclass[sigconf]{acmart}
\AtBeginDocument{%
  \providecommand\BibTeX{{%
    \normalfont B\kern-0.5em{\scshape i\kern-0.25em b}\kern-0.8em\TeX}}}

\copyrightyear{2021} 
\acmYear{2021} 
\setcopyright{acmcopyright}\acmConference[SIGIR '21]{Proceedings of the 44th International ACM SIGIR Conference on Research and Development in Information Retrieval}{July 11--15, 2021}{Virtual Event, Canada}

\acmBooktitle{Proceedings of the 44th International ACM SIGIR Conference on Research and Development in Information Retrieval (SIGIR '21), July 11--15, 2021, Virtual Event, Canada}
\acmPrice{15.00}
\acmDOI{10.1145/3404835.3463085}
\acmISBN{978-1-4503-8037-9/21/07}

\acmSubmissionID{sp1628}

\usepackage{enumitem}

\newcommand{\eg}{\textit{e}.\textit{g}.}

\begin{document}
\fancyhead{}
\title{Synthetic Target Domain Supervision for Open Retrieval QA}




\author{Revanth Gangi Reddy}
\email{revanth3@illinois.edu}
\affiliation{
\institution{UIUC}
\country{United States}
}
\author{Bhavani Iyer}
\email{bsiyer@us.ibm.com}
\affiliation{\institution{IBM Research AI}
\country{United States}}
\author{Md Arafat Sultan}
\email{arafat.sultan@ibm.com}
\affiliation{\institution{IBM Research AI}
\country{United States}}
\author{Rong Zhang}
\email{zhangr@us.ibm.com}
\affiliation{\institution{IBM Research AI}
\country{United States}}
\author{Avirup Sil}
\email{avi@us.ibm.com}
\affiliation{\institution{IBM Research AI}
\country{United States}}
\author{Vittorio Castelli}
\email{vittorio@us.ibm.com}
\affiliation{\institution{IBM Research AI}
\country{United States}}
\author{Radu Florian}
\email{raduf@us.ibm.com}
\affiliation{\institution{IBM Research AI}
\country{United States}}
\author{Salim Roukos}
\email{roukos@us.ibm.com}
\affiliation{\institution{IBM Research AI}
\country{United States}}
\renewcommand{\shortauthors}{Reddy, et al.}

\begin{abstract}
Neural passage retrieval is a new and promising approach in open retrieval question answering.
In this work, we stress-test the Dense Passage Retriever (DPR)---a state-of-the-art (SOTA) open domain neural retrieval model---on closed and specialized target domains such as COVID-19, and find that it lags behind standard BM25 in this important real-world setting.
To make DPR more robust under domain shift, we explore its fine-tuning with synthetic training examples, which we generate from unlabeled target domain text using a text-to-text generator.
In our experiments, this noisy but fully automated target domain supervision gives DPR a sizable advantage over BM25 in out-of-domain settings, making it a more viable model in practice.
Finally, an ensemble of BM25 and our improved DPR model yields the best results, further pushing the SOTA for open retrieval QA on multiple out-of-domain test sets.
\end{abstract}




\maketitle

\section{Introduction}

\begin{table*}[ht]
    \centering
    \small
    \begin{tabular}{p{9.3cm}|p{7.7cm}}
    \multicolumn{1}{c|}{\textbf{Passage}}     &  \multicolumn{1}{c}{\textbf{Synthetic Question-Answer pairs}}   \\
    \hline
    ... Since December 2019, when the first patient with a confirmed case of COVID-19 was reported in Wuhan, China, over 1,000,000 patients with confirmed cases have been reported worldwide. It has been reported that the most common symptoms include fever, fatigue, dry cough, anorexia, and dyspnea. Meanwhile, less common symptoms are nasal congestion ...
    & 
    \textbf{Q:} What are the most common symptoms of COVID-19?
         
        \textbf{A:} fever, fatigue, dry cough, anorexia, and dyspnea
        
        ~
        
        \textbf{Q:} How many people have been diagnosed with COVID-19?
        
        \textbf{A:} over 1,000,000\\
        \hline
         
        
     
        
    \end{tabular}
    \caption{Synthetic MRC examples generated by our generator from a snippet in the CORD-19 collection.}
    \label{tab:Synth-Ex}
    \vspace{-1em}
\end{table*}

Open retrieval question answering (ORQA) finds a short answer to a natural language question in a large document collection \cite{chen2017reading, lee2019latent,asai2020xor}.
Most ORQA systems employ (i) an information retrieval (IR) component that retrieves relevant passages from the given corpus \cite{seo2019real,lee2019latent,guu2020realm} and (ii) a machine reading comprehension (MRC) component that extracts the final short answer from a retrieved passage \cite{rajpurkar2016squad, alberti2019bert, liu2020rikinet}.
Recent work on ORQA by \citet{karpukhin2020dense} shows that distant supervision for neural passage retrieval can be derived from annotated MRC data, yielding a superior approach \cite{lewis2020retrieval, izacard2020leveraging} to classical term matching methods like BM25 \cite{chen2017reading,robertson2009probabilistic}.
Concurrent advances in tools like FAISS \cite{johnson2019billion} that support efficient similarity search in dense vector spaces have also made this approach practical: 
when queried on an index with 21 million passages, FAISS processes 995 questions per second (qps).
BM25 processes 23.7 qps per CPU thread in a similar setting \cite{karpukhin2020dense}.

Crucially, all training and test instances for the Dense Passage Retrieval (DPR) model in \cite{karpukhin2020dense} were derived from open domain Wikipedia articles.
This is a rather limited experimental setting, as many real-world ORQA use cases involve distant target domains with highly specialized content and terminology, for which there is no labeled data.
On \mbox{COVID-19}, for example, a large body of scientific text is available~\cite{wang2020cord}, but practically no annotated QA data for model supervision.\footnote{The handful of existing COVID-19 QA datasets \cite{mollercovid, tang2020rapidly, lee2020answering} are quite small in size and can only be used for evaluation.}
In this paper, we closely examine neural IR---DPR to be specific---in out-of-domain ORQA settings, where we find that its advantage over BM25 diminishes or disappears altogether in the absence of target domain supervision.

Domain adaptation is an active area of investigation in supervised learning; existing techniques for different target scenarios include instance weighting \cite{jiang2007instance}, training data selection using reinforcement learning \cite{liu2019reinforced} and transfer learning from open domain datasets \cite{wiese2017neural}. 
For pre-trained language models, fine-tuning on unlabeled target domain text has also been found to be a useful intermediate step \cite{gururangan-etal-2020-dont,zhang2020multi}, \eg, with scientific \cite{beltagy2019scibert} and biomedical text \cite{lee2020biobert, alsentzer2019publicly}.
To address the performance degradation of DPR in low-resource out-of-domain settings, we explore another approach: fine-tuning with synthetically generated examples in the target domain.
Our example generator is trained using open domain (Wikipedia) MRC examples \cite{rajpurkar2016squad}. It is then applied to target domain (biomedical) documents to generate synthetic training data for both retrieval and MRC.
Despite being trained on generic open domain annotations, our generator
yields target domain examples that significantly boost results in those domains.
It should be noted here that unlike most existing work in the QA literature where human annotated training examples are used to fine-tune a synthetically pre-trained model \cite{dhingra2018simple, alberti2019synthetic,unilm, sultan-etal-2020-importance}, we rely on only synthetic examples in the target domain.

The contributions of this paper are as follows:
\begin{itemize}
\item We empirically show the limitations of open domain
neural IR (DPR) when applied zero shot to ORQA in distant target domains.
\item We present a solution to this problem that relies on automatic text-to-text generation to create target domain synthetic training data. Our synthetic examples improve \textit{both IR and end-to-end ORQA results}, in \textit{both original and related target domains}, requiring \textit{no supervision with human annotated examples}.
\item We also show that ensembling over BM25 and our improved neural IR model
yields the best results---which underscores the complementary nature of the two approaches---further pushing the state of the art for out-of-domain ORQA on multiple benchmarks. 
\end{itemize}




\section{Method}
This section describes our methods for generating synthetic examples in the target domain and their application to both IR and MRC to construct the final ORQA pipeline.

\subsection{Generating Synthetic Training Examples}
Let $(p, q, a)$ be an MRC example comprising a passage $p$, a question $q$, and its short answer $a$ in $p$.
Let $s$ be the sentence in $p$ that contains the answer $a$.
In what follows, we train an example generator to produce the triple $(s, a, q)$ given $p$.
The answer sentence $s$ is subsequently used to locate $a$ in $p$, as a short answer text (\eg, a named entity) can generally occur more than once in a passage.

To train the generator, we fine-tune BART~\cite{lewis-etal-2020-bart}---a pre-trained denoising sequence-to-sequence generation model---with MRC examples from open domain datasets like SQuAD~\cite{rajpurkar2016squad}.
The generator $g$ with parameters $\theta_g$ learns to maximize the conditional joint probability $P(s,a,q|p;\theta_g)$.
In practice, we (i) only output the first ($s_f$) and the last ($s_l$) word of $s$ instead of the entire sentence for efficiency, and (ii) use special separator tokens to mark the three items in the generated triple.

Given a target domain passage $p$ at inference time, an ordered sequence $(s_f, s_l, [SEP], a, [SEP], q)$ is sampled from $g$ using top-$k$ top-$p$ sampling~\cite{Holtzman:2019}, which has been shown to yield better training examples than greedy or beam search decoding due to greater sample diversity~\cite{sultan-etal-2020-importance}.
From this generated sequence, we create positive synthetic training examples for both passage retrieval: $(q, p)$ and MRC: $(p, q, a)$, where $s_f$ and $s_l$ are used to locate $a$ in $p$.
Table~\ref{tab:Synth-Ex} shows two examples generated by our generator from a passage in the CORD-19 collection~\cite{wang2020cord}.

\subsection{Passage Retrieval}
\label{sec:ir}
As stated before, we use DPR~\cite{karpukhin2020dense} as our base retrieval model.
While other competitive methods such as ColBERT~\cite{khattab2020colbert} exist, DPR offers a number of advantages in real-time settings as well as in our target scenario where retrieval is only a component in a larger ORQA pipeline.
For example, by compressing each passage down to a single vector representation, DPR can operate with significantly less memory.
It is also a faster model for several reasons, including not having a separate re-ranking module.

For target domain supervision of DPR, we fine-tune its off-the-shelf open domain instance with synthetic examples.
At each iteration, a set of questions is randomly sampled from the generated dataset. 
Following \citet{karpukhin2020dense}, we also use in-batch negatives for training. 
We refer the reader to their article for details on DPR supervision. 
We call this final model the \emph{Adapted DPR} model.

\subsection{Machine Reading Comprehension}
For MRC, we adopt the now standard approach of ~\citet{devlin2019bert} that (i) starts from a pre-trained transformer language model (LM), (ii) adds two pointer networks atop the final transformer layer to predict the start and end positions of the answer phrase, and (iii) fine-tunes the entire network with annotated MRC examples.
We choose RoBERTa~\cite{liu2019roberta} as our base LM.
Given our out-of-domain target setting, we fine-tune it in two stages as follows.

First, the RoBERTa LM is fine-tuned on unlabeled target domain documents, which is known to be a useful intermediate fine-tuning step~\cite{gururangan-etal-2020-dont}.
This target domain model is then further fine-tuned for MRC, where we use both human annotated open domain MRC examples and target domain synthetic examples, as detailed in Section~\ref{sec:experiments}.
Additionally, we denoise the synthetic training examples using a roundtrip consistency~\cite{alberti2019synthetic} filter: an example is filtered out if its candidate answer score, obtained using an MRC model trained on SQuAD 2.0 and NQ, is lower than a threshold $t$ ($t$ tuned on a validation set).

\begin{table*}[ht!]
\centering
\begin{tabular}{l||ccc|ccc||ccc}
\multicolumn{1}{c||}{Model} &  \multicolumn{6}{c||}{Open-COVID-QA-2019} & \multicolumn{3}{c}{COVID-QA-111} \\
\hline

\multicolumn{1}{c||}{} &  \multicolumn{3}{c|}{Dev} & \multicolumn{3}{c||}{Test} & \multicolumn{3}{c}{Test} \\
\multicolumn{1}{c||}{}  & M@20 & M@40 & M@100 & M@20 & M@40 & M@100 & M@20 & M@40 & M@100                         \\ 
                                                           
\hline
BM25 & 22.4 & 24.9 &  29.9&  29.9&  33.4&  39.7&  48.7& 60.4 & 64.9\\
DPR-Multi  &  14.4&  18.4&  22.9&  13.8&  17.5&  21.4& 51.4 &  57.7& 66.7\\
ICT & 16.6 & 21.6 & 25.5 & 18.1 & 23.0 & 29.6 & 52.8 & 59.8 & 67.6 \\
\hline
Adapted DPR & 28.0 & 31.8 & 39.0 & 34.8 &  40.4 &  47.2 &  58.6 & 64.6 & 74.2 \\
\hline
BM25 + DPR-Multi &  23.4&  27.9&  32.3&  29.5&  33.2&  38.9& 58.6 & 65.8 & 69.4 \\
BM25 + Adapted DPR & \textbf{31.8} & \textbf{36.0} & \textbf{42.6} & \textbf{43.2} & \textbf{48.2} & \textbf{53.7} & \textbf{60.4} & \textbf{68.2} & \textbf{76.9} \\
\hline
\end{tabular}
\caption{Performance of different IR systems on (a) the open retrieval version of COVID-QA-2019, and (b) COVID-QA-111.} 
\label{tab:IR}
\vspace{-1em}
\end{table*}

\subsection{Open Retrieval Question Answering}
Using the described retrieval and MRC components, we construct our final ORQA system that executes a four-step process at inference time.
First, only the $K$ highest scoring passages returned by IR for the input question are retained ($K$ tuned on a validation set).
Each passage is then passed along with the question to the MRC component, which returns the respective top answer and its MRC score.
At this point, each answer has two scores associated with it: its MRC score and the IR score of its passage.
In the third step, these two scores get normalized using the Frobenius norm and combined using a convex combination.
The weight in the combination operation is tuned on a validation set.
Finally, the answer with the highest combined score is returned.







\section{Experimental Setup}
\label{sec:experiments}
We evaluate the proposed systems on out-of-domain retrieval, MRC, and end-to-end ORQA against SOTA open domain baselines.

\subsection{Retrieval Corpus and Datasets}
\label{sec:datasets}
We select COVID-19 as our primary target domain, an area of critical interest at the point of the writing.
We use 74,059 full text PDFs from the June 22, 2020 version of  CORD-19~\cite{wang2020cord} document collection on SARS-CoV-2--and related coronaviruses as our retrieval corpus.
Each document is split into passages that (a)~contain no more than 120 words, and (b)~align with sentence boundaries, yielding around 3.5 million passages.

We utilize three existing datasets for COVID-19 target domain evaluation.
The first one, used to evaluate  retrieval and MRC results separately as well as end-to-end ORQA performance, is \textit{COVID-QA-2019}
\cite{mollercovid}---a dataset of question-passage-answer triples created from COVID-19 scientific articles by volunteer biomedical experts. 
We split the examples into Dev and Test subsets of 203 and 1,816, respectively.
Since end-to-end ORQA examples consist of only question-answer pairs with no passage alignments, we also create a version of this dataset for ORQA evaluation (\textit{Open-COVID-QA-2019} henceforth) wherein duplicate questions are de-duplicated and different answers to the same question are all included in the set of correct answers, leaving 201 Dev and 1,775 Test examples.

Our second dataset---\textit{COVID-QA-147}
\cite{tang2020rapidly}---is a QA dataset obtained from Kaggle's CORD-19 challenge,
containing 147 question-article-answer triples with 27 unique questions and 104 unique articles. Due to the small number of unique questions in this dataset, we only use it for out-of-domain MRC  evaluation.

Finally, \textit{COVID-QA-111}
\cite{lee2020answering} contains queries gathered from different sources, \eg, Kaggle and the FAQ sections of the CDC
and the WHO.
It has 111 question-answer pairs with 53 interrogative and 58 keyword-style queries. Since questions are not aligned to passages in this dataset, we use it only to evaluate IR and ORQA.


\subsection{Synthetic Example Generation}
\label{sec:setup-synthgen}
We fine-tune BART for three epochs on the open domain MRC training examples of SQuAD1.1 \cite{rajpurkar2016squad} (lr=3e-5).
Synthetic training examples are then generated for COVID-19  from the CORD-19 collection. 
We split the articles into chunks of at most 288 wordpieces and generate five MRC examples from each of the resulting 1.8 million passages. 
For \mbox{top-$k$} \mbox{top-$p$} sampling, we use $k$=$10$ and $p$=$0.95$.
Overall, the model generates about 7.9 million examples.

\subsection{Retrieval and MRC}
We use the DPR-Multi system
from \cite{karpukhin2020dense}
as our primary neural IR baseline. DPR-Multi comes pre-trained on open-retrieval versions of several MRC datasets: Natural Questions (NQ)~\cite{kwiatkowski2019natural}, WebQuestions~\cite{berant2013semantic}, CuratedTrec~\cite{baudivs2015modeling} and TriviaQA~\cite{joshi2017triviaqa}. 
We fine-tune it for six epochs with COVID-19 synthetic examples to train our \textit{Adapted DPR} model (lr=1e-5, batch size=128).
We also evaluate the Inverse Cloze Task (ICT) method~\cite{lee2019latent} as a second neural baseline, which masks out a sentence at random from a passage and uses it as a query to create a query-passage training pair. 
We use ICT to fine-tune DPR-Multi on the CORD-19 passages of Section~\ref{sec:setup-synthgen}, which makes it also a synthetic domain adaptation baseline.
Finally, for each neural IR model, we also evaluate its ensemble with BM25 that computes a convex combination of normalized neural and BM25 scores.
The weight for BM25 in this combination is 0.3 (tuned on Open-COVID-QA-2019 Dev).

Our baseline MRC model is based on a pre-trained RoBERTa-Large LM,  and is fine-tuned for three epochs on SQuAD2.0 and then for one epoch on NQ. 
It achieves a short answer EM of 59.4 on the NQ dev set, which is competitive with numbers reported in \cite{liu2020rikinet}. 
For target domain training, we first fine-tune a RoBERTa-Large LM on approximately 1.5GB of CORD-19 text containing 225 million tokens (for 8 epochs, lr=1.5e-4).
The resulting model is then fine-tuned for MRC for three epochs on SQuAD2.0 examples and one epoch each on roundtrip-consistent synthetic MRC examples and NQ. 
For roundtrip consistency check, we use a threshold of $t$=$7.0$, which leaves around 380k synthetic examples after filtering. 

\begin{table*}[t]
\centering
\begin{tabular}{l||cc|cc||cc}
\multicolumn{1}{c||}{Model} & \multicolumn{4}{c||}{Open-COVID-QA-2019} & \multicolumn{2}{c}{COVID-QA-111} \\
\hline
\multicolumn{1}{c||}{} & \multicolumn{2}{c|}{Dev} & \multicolumn{2}{c||}{Test} & \multicolumn{2}{c}{Test} \\
\multicolumn{1}{c||}{} & Top-1 & Top-5 & Top-1 & Top-5 & Top-1 & Top-5 \\
\hline
BM25 $\rightarrow$ Baseline MRC & 21.7 & 31.8 & 27.1 & 38.7  & 24.1 & 39.3\\
(BM25 + DPR-Multi) $\rightarrow$ Baseline MRC & 21.4 & 30.9 & 25.2 & 37.2 & 24.4 & 43.2 \\
(BM25 + Adapted DPR) $\rightarrow$ Baseline MRC & 24.2 & 35.6 & 29.5 & 44.2  & 25.0 & 45.9 \\
(BM25 + Adapted DPR) $\rightarrow$ Adapted MRC & \textbf{27.2} & \textbf{37.2} &  \textbf{30.4} & \textbf{44.9} & \textbf{26.5} & \textbf{47.8} \\
\hline
\end{tabular}
\caption{End-to-end F1 scores achieved by different Open retrieval QA systems.
The best system (last row) utilizes target domain synthetic training examples for both IR and MRC supervision.} 
\label{tab:Open_QA}
\vspace{-1em}
\end{table*}

\subsection{Metrics}
We evaluate IR using  Match@$k$, for $k \in \{20,40,100\}$ \cite{karpukhin2020dense}.
For MRC, we use standard Exact Match (EM) and F1 score.
Finally, end-to-end ORQA accuracy is measured using Top-1 and Top-5 F1.

\section{Results and Analysis}

We first report results separately for IR and MRC. 
Then we evaluate ORQA pipelines that must find a short answer to the input question in the CORD-19 collection.
Reported numbers for all trained models are averages over three random seeds.


\subsection{Passage Retrieval}

Table \ref{tab:IR} shows performances of different IR systems on Open-COVID-QA-2019 and COVID-QA-111.
BM25\footnote{\href{https://lucene.apache.org}{Lucene} Implementation. BM25 parameters $b=0.75$ (document length normalization) and $k_1=1.2$ (term frequency scaling) worked best.} demonstrates robust results relative to the neural baselines. 
While DPR-multi is competitive with BM25 on COVID-QA-111, it is considerably behind on the larger Open-COVID-QA-2019. 
ICT improves over DPR-multi, indicating that even weak target domain supervision is useful.
The proposed Adapted DPR system achieves the best single system results on both datasets, with more than 100\% improvement over DPR-Multi on the Open-COVID-QA-2019 Test set.
Finally, ensembling over BM25 and neural approaches yields the best results. 
The BM25+Adapted DPR ensemble is the top system across the board, with a difference of at least 14 points with the best baseline on the Open-COVID-QA-2019 Test set (all metrics), and 8 points on COVID-QA-111.

Upon closer examination, we find that BM25 and Adapted DPR retrieve passages that are very different. 
For Open-COVID-QA-2019, for example, only 5 passages are in common on average between the top 100 retrieved by the two systems. 
This diversity in retrieval results explains why they complement each other well in an ensemble system, leading to improved IR performance.

\subsection{Machine Reading Comprehension}
\label{sec:results-mrc}
Table \ref{tab:QA} shows results on the two COVID-19 MRC datasets. 
Input to each model is a question and an annotated document that contains an answer. 
Our proposed model achieves 2.0--3.7 F1 improvements on the Test sets over a SOTA open domain MRC baseline.
On the COVID-QA-2019 Dev set, we see incremental gains from applying the two domain adaptation strategies.

\begin{table}[h]
\centering
\small
\vspace{-2.3mm}
\begin{tabular}{l||cc|cc||cc}
\multicolumn{1}{c||}{Model} & \multicolumn{4}{c||}{COVID-QA-2019} & \multicolumn{2}{c}{COVID-QA-147} \\ 
\hline
\multicolumn{1}{c||}{} & \multicolumn{2}{c|}{Dev} & \multicolumn{2}{c||}{Test} & \multicolumn{2}{c}{Test} \\
\multicolumn{1}{c||}{} & EM & F1 & EM & F1 &  EM & F1                          \\ 
                                                           
\hline
Baseline MRC  & 34.0 & 59.4 & 34.7 &  62.7 & 8.8 & 31.0\\
+ CORD-19 LM & 35.5 & 60.2 & - & - & - & - \\
+ Syn. MRC training   & \textbf{38.6} & \textbf{62.8} & \textbf{37.2} &  \textbf{64.7} & \textbf{11.3} & \textbf{34.7}\\
\hline
\end{tabular}
\caption{MRC performances on COVID-19 datasets. The last row refers to the proposed model that is trained on unlabeled CORD-19 text as well as synthetic MRC examples. } 
\label{tab:QA}
\vspace{-2em}
\end{table}



\subsection{Open Retrieval Question Answering}
Using different pairings of the above IR and MRC systems, we build several ORQA pipelines.
Each computes a convex combination of its component IR and MRC scores after normalization, with the IR weight being 0.7 (tuned on Open-COVID-QA-2019 Dev).
We observe that retrieving $K$=100 passages is optimal when IR is BM25 only, while $K$=40 works best for BM25+Neural IR.

Table~\ref{tab:Open_QA} shows end-to-end F1 scores of the different ORQA pipelines.
Adapted MRC refers to the best system of Section~\ref{sec:results-mrc} (Table~\ref{tab:QA} Row 3).
Crucially, the best system in Table~\ref{tab:Open_QA} (last row) uses synthetic target domain supervision for both IR and MRC.
In a paired $t$-test~\cite{hsu2005paired} of the Top-5 F1 scores, we find the differences with the baseline (Row 1) to be statistically significant at $p$$<$$0.01$.

\subsection{Zero Shot Evaluation on BioASQ}
To investigate if our synthetically fine-tuned COVID-19 models can also help improve performance in a related target domain, we evaluate them zero shot on the BioASQ~\cite{balikas2015bioasq} task. 
BioASQ contains biomedical questions with answers in the PubMed
abstracts.
For evaluation, we use the factoid questions from the Task 8b
training and test sets, totaling  1,092 test questions.
As our retrieval corpus, we use around 15M abstracts from Task 8a. 
We pre-process them as described in Section \ref{sec:datasets} to end up with around 37.4M passages.

\begin{table}[h]
\centering
\begin{tabular}{lccc}
\multicolumn{1}{c}{Model} & M@20 & M@40 & M@100 \\ 
\hline
BM25 & 42.1 & 46.4 & 50.5 \\
DPR-Multi & 37.6 & 42.8 & 48.1 \\
Adapted DPR & \textbf{42.4} & \textbf{48.9} & \textbf{55.9} \\
\hline
\end{tabular}
\caption{IR results on BioASQ Task 8B factoid questions.} 
\label{tab:Bioasq_IR}
\vspace{-1.5em}
\end{table}

Table~\ref{tab:Bioasq_IR} shows the BioASQ retrieval results, where the proposed Adapted DPR model outperforms both baselines. 
Table~\ref{tab:Bioasq_end_to_end} summarizes the evaluation on the end-to-end ORQA task, where we see similar gains from synthetic training. 
These results show that synthetic training on the CORD-19 articles transfers well to the broader related domain of biomedical QA. 

\begin{table}[h]
\centering
\begin{tabular}{lccc}
\multicolumn{1}{c}{Model} & Top-1 & Top-5 \\ 
\hline
BM25 $\rightarrow$ Baseline MRC & 30.6 & 45.5\\
DPR-Multi $\rightarrow$ Baseline MRC & 28.6 & 43.0\\
Adapted DPR $\rightarrow$ Baseline MRC & 32.1 & 49.4\\
Adapted DPR $\rightarrow$ Adapted MRC & \textbf{32.9} & \textbf{49.5}\\
\hline
\end{tabular}
\caption{ORQA F1 scores on BioASQ 8B factoid questions.} 
\label{tab:Bioasq_end_to_end}
\vspace{-2em}
\end{table}

\section{Conclusion}
Low-resource target domains can present significant challenges for supervised language processing systems.
In this paper, we show that synthetically generated target domain examples can support strong domain adaptation of neural open domain open retrieval QA models, which can further generalize to related target domains.
Crucially, we assume zero labeled data in the target domain and rely only on open domain MRC annotations to train our generator.
Future work will explore semi-supervised and active learning approaches to examine if further improvements are possible with a small amount of target domain annotations.

\bibliographystyle{ACM-Reference-Format}
\bibliography{sample-base}










\end{document}